# On-Chip Learning with Memristor-Based Neural Networks: Assessing Accuracy and Efficiency Under Device Variations, Conductance Errors, and Input Noise

M.Reza Eslami,[1] Dhiman Biswas,[2] Soheib Takhtardeshir,[3] Sarah S. Sharif,[4] and Yaser M. Banad[4]

[1] Electrical & Computer Engineering Department, Florida State University (FSU), Tallahassee, FL 32310 USA
[2] Department of Physics and Astronomy, The University of Oklahoma, Norman, OK 73019, USA
[3] Department of Computer and Electrical Engineering, Mid Sweden University, Sundsvall, SE 85170, Sweden
[4] School of Electrical and Computer Engineering, The University of Oklahoma, Norman, OK 73019, USA
Corresponding author: bana@ou.edu

*Abstract*— This paper presents a memristor-based compute-in-memory hardware accelerator for on-chip training and inference, focusing on its accuracy and efficiency against device variations, conductance errors, and input noise. Utilizing realistic SPICE models of commercially available silver-based metal self-directed channel (M-SDC) memristors, the study incorporates inherent device non-idealities into the circuit simulations. The hardware, consisting of 30 memristors and 4 neurons, utilizes three different M-SDC structures with tungsten, chromium, and carbon media to perform binary image classification tasks. An on-chip training algorithm precisely tunes memristor conductance to achieve target weights. Results show that incorporating moderate noise (<15%) during training enhances robustness to device variations and noisy input data, achieving up to 97% accuracy despite conductance variations and input noises. The network tolerates a 10% conductance error without significant accuracy loss. Notably, omitting the initial memristor reset pulse during training considerably reduces training time and energy consumption. The hardware designed with chromium-based memristors exhibits superior performance, achieving a training time of 2.4 seconds and an energy consumption of 18.9 mJ. This research provides insights for developing robust and energy-efficient memristor-based neural networks for on-chip learning in edge applications.

## I. Introduction

Memristors, first conceptualized by Leon Chua in 1971 [1] and physically realized by Hewlett-Packard in 2008 [2] represents a fundamental circuit element that links electric charge and magnetic flux.
These devices have garnered significant attention for diverse applications due to their advantageous properties, such as low power consumption [3-5], rapid switching speeds [6], high endurance [7, 8], exceptional scalability [9], and compatibility with CMOS technology [10-12]. Beyond their promise for memory and data storage, memristors hold significant potential for unconventional computing, particularly in neuromorphic applications where they can emulate the behavior of biological synapses [13-17]. Memristors' unique ability to co-locate storage and computation within the same device enables in-memory computing, eliminating the data movement bottleneck inherent in traditional von Neumann architectures. This capability, coupled with their non-volatility, allows for highly parallel and energy-efficient execution of matrix-vector multiplications [18, 19], the fundamental operations to accelerate computations in neural networks (NNs) [20-24].

While memristor-based neural networks hold promise for image classification [25, 26], many studies rely on software-only training procedures [27-29], often overlooking the performance degradation introduced by device non-idealities [30-32], during on-chip training and inference. This gap hinders a comprehensive understanding of the true energy efficiency and accuracy achievable in practical memristor-based image classification systems. Such performance measures enable on-chip training to utilize the dynamic trainability of memristors within the full analog range to directly accommodate device variations. However, existing research on on-chip training of memristor-based NNs has primarily focused on specific learning rules [30, 33, 34], or simple network architectures [35-37]. This limited scope hinders the exploration of more generalizable on-chip training methods and a thorough understanding of how device non-idealities impact accuracy and efficiency in complex tasks. While some studies have explored on-chip training of memristor-based NNs [38], a comprehensive understanding of how device non-idealities like device-to-device (D2D) variations, conductance errors, and noise impact the accuracy and efficiency of these systems remains limited [39].

This paper presents a comprehensive circuit design for a memristor-based neural network hardware accelerator, specifically tailored for on-chip training and inference. Our approach uniquely incorporates realistic memristor models derived from experimental data of the first commercialized memristors [40], capturing inherent memristor non-idealities. This design, implemented within the Proteus circuit simulation environment, integrates both a custom memristor component and a microcontroller to enable on-chip training and inference procedures. Furthermore, our work evaluates the performance of the memristor-based neural network in classifying noisy input images, a crucial aspect often overlooked in studies that rely on idealized simulations with pristine input data [41, 42].

.

This approach bridges a gap in existing research, which often relies on idealized simulations, input signals, or external measurement equipment, allowing us to explore the true potential and challenges of memristor-based neuromorphic computing for complex classification tasks in edge applications.

The structure of this paper is organized as follows: Section II, focuses on representing the characteristics of real memristors. We analyze the I-V behaviors of silver-based Metal Self-Directed Channel (M-SDC) memristors [40], evaluating 16 discrete devices from three M-SDC structures (tungsten, chromium, and carbon). We then develop SPICE models for their non-ideal I-V curves and, integrate these models into the Proteus environment's library for the first time, enabling comprehensive circuit simulations with realistic memristor behavior. Section III introduces a two-layer neural network architecture tailored for hardware implementation and image classification tasks. This network serves as the foundation for our memristor-based accelerator and is designed to classify 3x3 pixel images. Section IV details the optimized circuit design of our hardware accelerator, encompassing 30 memristors, 4 neurons, and peripheral circuitry essential for on-chip training and inference, providing a detailed blueprint for practical realization. Section V explains the training algorithm, executed by a microcontroller, that precisely adjusts the electrical conductivity of each memristor within the neural network circuit. This algorithm, a core element of our on-chip training system, ensures accurate and efficient weight updates during the training phase. Section VI presents a rigorous evaluation of our proposed system through multiple analyses. We compare the accuracy of the hardware-implemented neural network with its numerical counterpart, assessing the impact of D2D variations and noise on classification performance. The network's generalizability is demonstrated through its capability to perform binary classification for various 9-pixel images. We further explore the impact of memristor conductance errors on network accuracy, determining tolerance limits and highlighting the importance of on-chip retraining. Finally, we analyze the time and energy consumption of both training and inference, revealing the efficiency advantages of different memristor types and the trade-offs associated with initial reset pulses. This comprehensive evaluation provides valuable insights into the performance, robustness, and efficiency of our proposed memristor-based neural network hardware accelerator.

## II. Memristor Modeling

### A. Exploring the I-V of Real Memristors

Fig. 1 showcases the material composition and representative I-V characteristics of three M-SDC memristors, each utilizing a distinct active medium: tungsten, chromium, or carbon. The devices share a common structure, comprising tungsten electrodes and a crucial silver layer for conduction, supported by a SnSe ion conductor that facilitates the controlled migration of silver ions to form conductive pathways [43].

Typically, I-V curves exhibit two prominent slopes: a steeper slope indicating the low resistive state (LRS) and a gentler slope corresponding to the high resistive state (HRS). The transition between these states is marked by numerous current jumps, each associated with different SET (transitioning to LRS) or RESET (transitioning to HRS) voltages. The variations in SET and RESET voltages across different memristors are attributed to the statistical nature of filament formation within them [6]. Factors such as the number and strength of filaments formed in each cycle contribute to distinct current jump patterns, impacting key memristor parameters like Von, Voff, Ron, and Roff. The active medium alters the dynamic filament formation and thereby the device's resistance state, resulting in distinct I-V characteristics. In the absence of connected filaments, the device remains in a high resistive state (HRS). Applying a forward voltage initiates the agglomeration of silver ions, encouraging the growth of Conducting Memristive Filaments (CMFs) [44], and ultimately transitioning the device into a low resistive state (LRS) upon complete filament formation.

To capture the realistic behavior of these memristors, we characterized 16 discrete devices from each of the three M-SDC structures. Their nominal I-V characteristics (blue curves) and individual I-V curves (gray lines) are shown in Fig.1, measured at 1 Hz using a sinusoidal voltage with an amplitude of 0.5V. This measurement falls well within the safe operating range of 0.1 to 0.75V for resistance modulation [40], ensuring that the devices are not damaged during characterization. Operating beyond this voltage range without proper current limitation risks device damage, highlighting the need for careful consideration during circuit design and operation.

### B. Incorporating the Non-Ideal I-V Characteristic of Real Memristors Within a SPICE Model

Despite undergoing a standardized fabrication process, memristors exhibit notable D2D variability, resulting in distinct I-V characteristics. This variability stems from inherent stochasticity in filament formation, as well as potential variations in material properties, linewidth roughness (LWR), and line edge roughness (LER) [45]. These factors significantly influence the electrical performance and reliability of memristors, leading to variations in resistance states, conductance levels, and ultimately, the accuracy and efficiency of memristor-based circuits.

To ensure realistic simulations, we selected one experimental I-V curves from each memristor type, capturing the non-ideal behavior typical of these devices (blue curves in Fig. 1). We then developed SPICE models (red curve in Fig. 1), meticulously tuning their parameters (provided in Appendix D) to replicate the observed experimental behavior. A detailed analysis of the influence of each model parameter is provided in Appendix B, and the corresponding SPICE model code is available in Appendix C.

### C. Developing the Memristor Component in Proteus Circuit Simulator

In a novel contribution, we integrated our developed SPICE memristor models into the Proteus software environment,

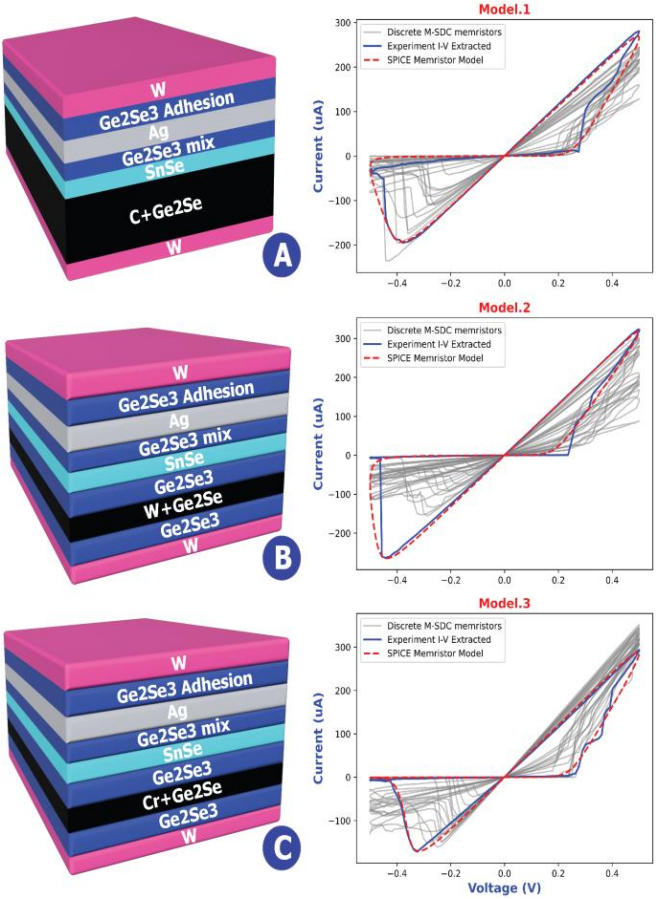

**Fig. 1.** Experimental and simulated I-V characteristics of three M-SDC memristor structures with active regions based on (A) carbon, (B) tungsten, and (C) chromium. The blue curves represent the nominal experimental I-V behaviors of the selected devices, while the gray lines show individual I-V curves from 16 devices of each type. The red curves illustrate the corresponding SPICE models developed to replicate the experimental behavior.

enabling the training and inferencing simulations of memristor-based neural network circuits. By incorporating our customized memristor alongside a vast library of commercially available components and microcontrollers, we achieve a more realistic and comprehensive representation of the hardware, improving the applicability of our findings to practical, inexpensive implementations. We designed a custom 32-pin memristor component, incorporating 16 discrete memristors with three different models, as illustrated in Fig. 12 in Appendix A. Details regarding the design and creation of this memristor element in Proteus, including the assignment of the SPICE subcircuit model and microcontroller code examples, are available from the corresponding author upon request.

### III. NETWORK ARCHITECTURE

To demonstrate the feasibility and assess the performance of our proposed memristor-based hardware accelerator, we focus on a two-layer artificial neural network (ANN) designed for binary classification tasks. This network, illustrated in Fig. 2, is specifically tailored for classifying all possible pairs of 3×3 pixel images, in this case, the standard English letters 'O' and 'X'. While this image size represents a simplified scenario, it allows for a comprehensive exploration of the network's capabilities and the impact of device non-idealities on accuracy and efficiency within a manageable computational scope.

The network model comprises nine binary inputs, corresponding to pixel values of 0 or 1 (V). The first layer consists of 27 synaptic memristors organized in a 9x3 crossbar array, each connected to a neuron with a sigmoid activation function. The second layer includes three synaptic memristors (3x1 array), leading to a single output neuron with a sigmoid activation function. This two-layer architecture provides sufficient complexity to capture non-linear decision boundaries while remaining computationally tractable for our hardware implementation.

To validate the functionality of our network structure and generate training and testing datasets, we first developed a Python model. This model utilizes the NumPy library to create a large number of sample images, including variations with random noise (1-5% error rate) to simulate potential input pixel inaccuracies. By applying these images to the network, we can evaluate its accuracy and robustness under realistic operating conditions. The test datasets and their corresponding labels are then converted into an integer numerical format for storage in the microcontroller's memory, facilitating the on-chip training and inference procedures.

### IV. DESIGNING CIRCUIT AND SIMULATION

Fig. 3 presents the simplified schematic of the memristor-based artificial neural network, and Fig. 12 in Appendix A details the complete circuit implementation, including peripheral components in the software simulator. The circuit implementation of neural networks in Fig. 2 comprises 30 memristors and 4 neurons, along with peripheral circuitry for control and reconfiguration. Eight MAX335 ICs, each containing eight digitally controlled analog switches, provide independent control over the memristor connections, effectively managing the numerous switches within the network. Additionally, 16 DG418 switches are employed to reconfigure the circuit for training and testing modes, including facilitating neuron connections during inference.

A Mega Arduino microcontroller orchestrates the entire process, controlling the switches, executing the training algorithm, and overseeing both training and inference procedures. This microcontroller integration enables the simulation and exploration of on-chip training and control using a readily available and widely used platform, enhancing the practical relevance of our design.

The neuron element, defined as a SPICE sub-circuit (Appendix E), implements a sigmoid activation function to emulate neuronal behavior. The sigmoid function, incorporating a scaling factor to accommodate the conductance range of the memristors and the input voltage levels, is given by:

$$V_{out} = \frac{1}{1 + e^{-(10^4 \cdot I_{in} + V_b)}} \quad (1)$$

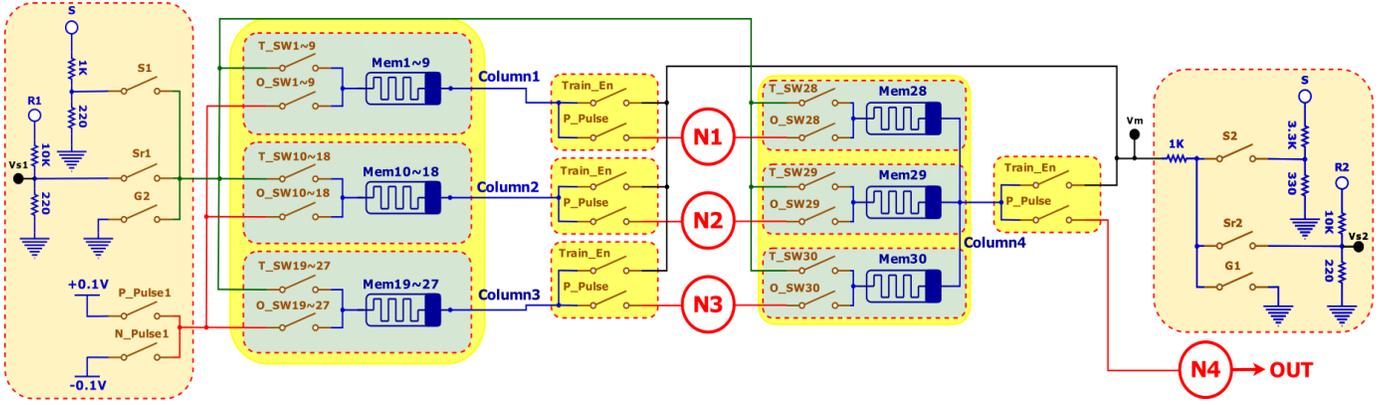

**Fig. 2.** Simplified schematic of the memristor-based neural network hardware accelerator, including 30 memristors and 4 neurons. The complete circuit implementation with peripheral components is detailed in Fig. 12 in Appendix A.

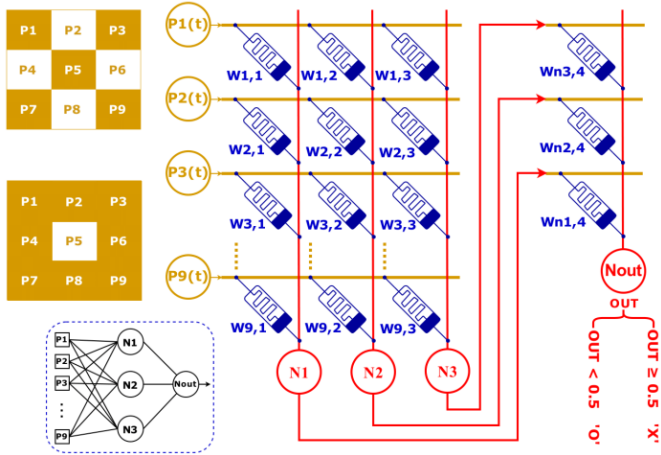

**Fig. 3.** Overview of the proposed two-layer memristor crossbar network, demonstrating the application of 3x3 pixel inputs representing the characters 'X' and 'O', and the corresponding output neuron states.

Where '$V_{out}$' is the neuron's output voltage, '$V_b$' is the neuron's bias voltage, and '$I_{in}$' is the sum of all memristors currents flowing through the memristors in a given column, calculated as:

$$I_{in} = \sum_{i=0}^{n-1} V_i \cdot G_i \qquad (2)$$

Here, '$V_i$' represents the voltage of input pixel '$i$', '$G_i$' is the conductance of memristor '$i$', and '$n$' is the number of memristors in the column. The scaling factor of '$10^4$' in Equation (1) accounts for the memristor conductances being in the millisiemens (mS) range and the input pixel voltages being 0.1V.

Recognizing the impracticality of negative weights in our hardware implementation, the training algorithm is designed to maintain all synaptic weights as positive values. Furthermore, the algorithm enforces a conductance range of 0 to 2.5 mS for each memristor, reflecting the limitations of attainable memristor conductance in practical devices.

### A. Training Circuit Configuration

To configure the circuit for memristor training, each memristor is individually selected via its 'T_SW' switch, connecting it to peripheral circuits responsible for pulse generation and training control as shown in Figures 4A and 4B. During the training phase, when a memristor is chosen, its 'T_SW' switch is activated by the microcontroller, connecting its positive terminal to the primary training pulse generation circuit (left side of the figure). The 'O_SW' and 'T_SW' switches of all other memristors remain inactive. Each column also has a 'Train_En' switch, activated during the training, which connects the negative terminals of that column to the secondary training pulse generation circuit (right side).

For memristor potentiation, the negative terminal is linked to the ground via a '1K' series resistor and the 'G1' switch, while the positive terminal is connected to the microcontroller's pulse output through the 'S1' switch (Fig. 4A). A voltage divider (1K and 220 Ohm) attenuates the '5V' microcontroller pulse to '1V' for forward biasing. For depreciation, the positive terminal is grounded via the 'G2' switch, and the negative terminal (through a 1K resistor) is connected to the pulse output via the 'S2' switch (Fig. 4B). Another voltage divider (3.3K and 330 Ohm) attenuates the pulse to '0.5V' for reverse biasing.

### B. Inferencing Circuit Configuration

Upon completion of training, the 'T_SW', 'G1', and 'G2' switches are deactivated, transitioning the circuit to the inference mode. Test data, stored in the microcontroller's memory, is sequentially applied to the network inputs ('P1-P9') over 100 cycles.

In each cycle, the 'O_SW' switches of memristors connected to input pixels with a value of '1' are activated. This connects their positive terminals to the 'P_Pulse1' and 'N_Pulse1' switches, and their negative terminals (through 'P_Pulse' switches) to the corresponding neuron as shown in Fig. 4C.

Each neuron receives a bias voltage determined during training. To read the network output, a 50ms positive pulse is applied via the 'P_Pulse1' switch, enabling measurement within the microcontroller's A/D converter cycle as shown in Fig. 5. An output neuron value greater than or equal to '0.5V' signifies recognition of 'X', while a value less than '0.5V' indicates 'O'. Fig. 5 also illustrates sample test data, including cases with one-pixel malfunctions, showing the network's classification performance.

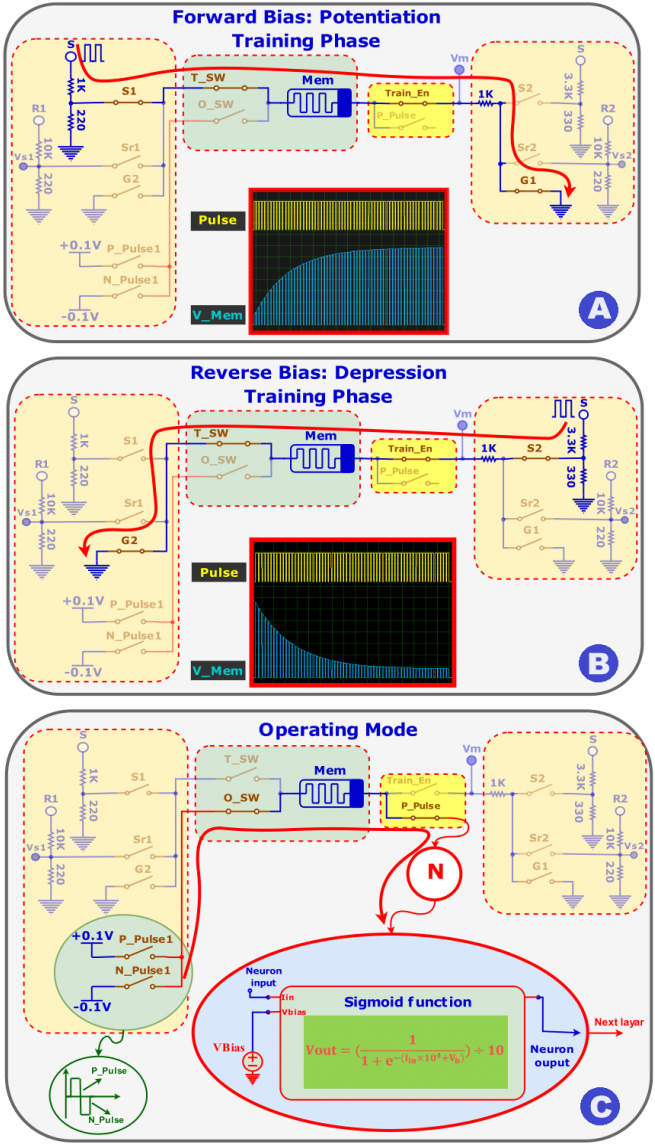

**Fig. 4.** Circuit configurations for (A) forward bias and potentiation and (B) reverse bias and depression of an individual memristor during the training phase. (C) Biasing configuration and reading pulse application during the inference phase, along with the neuron circuit.

The network's accuracy is evaluated by comparing the output value with the input image's label. If the classification is correct, an "Accuracy" variable is incremented. A 50ms negative pulse is then applied to restore memristor conductance before the next test data is applied. This process repeats for all 100 test data points, determining the overall network accuracy.

## V. TRAINING AND INFERENCING ALGORITHM

An optimal memristor training algorithm, implementable on memristor-based circuits, is crucial for advancing neuromorphic computing, enabling faster convergence, improved accuracy, and reduced energy consumption for more sophisticated and adaptable AI systems. This work presents such an algorithm, illustrated in the flowchart in Fig. 6, for on-chip training of memristors and adjusting synaptic weights. The training algorithm begins by examining the weights extracted from the trained neural network model to train the 30 synaptic memristors. For each memristor with a non-zero weight, the training operation commences after resetting its electrical conductivity (typically zero) using a reset pulse. This initialization step is crucial for ensuring accurate and consistent network performance by eliminating residual effects from previous operations [21].

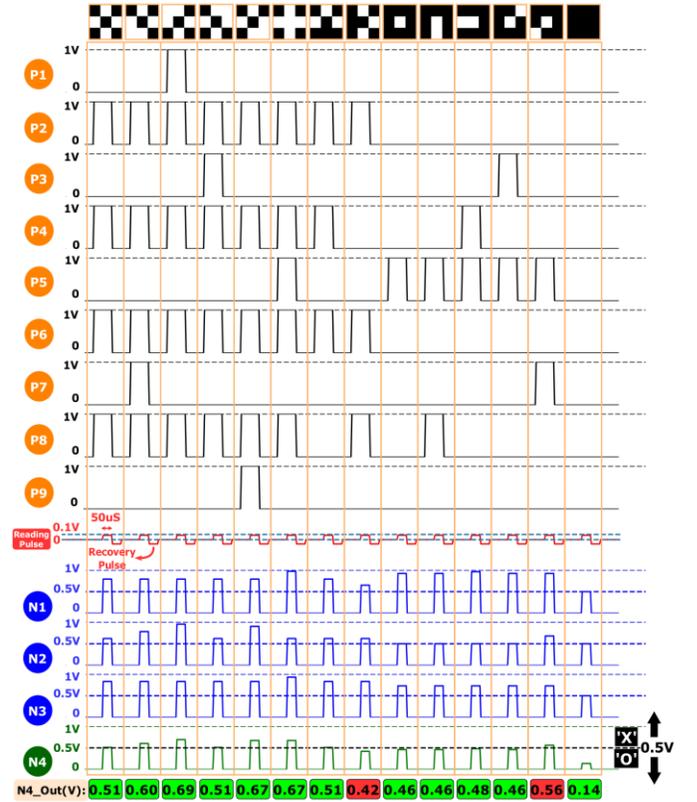

**Fig. 5.** Input pixels, reading pulses, and neuron output voltages for sample test data. The top row shows examples of input characters ('X' and 'O') with and without one-pixel malfunctions (noise). The subsequent rows illustrate the corresponding input pixel activations, reading pulses applied to the output neuron, and the resulting output neuron voltages. Red bars at the bottom indicate misclassifications, while green bars represent correct classifications.

To assess the impact of this initial reset, we conducted a comparative study (Fig. 7) examining the number of training cycles required to reach the desired weight for all memristors with non-zero weights, both with and without the reset pulse. This comparison was performed for each of the three memristor models (tungsten, chromium, and carbon). While there was minimal difference in most cases, Figures 7B and 7E reveal that certain weight adjustments might require significantly more cycles without the reset pulse (see orange and blue memristor curves) This discrepancy, potentially leading to higher delay and energy consumption, is further analyzed in the Results section (Fig. 11). While resetting memristors can reduce the total number of training cycles, the continuous application of a 200ms reset pulse to all memristors at the beginning of training introduces an overhead in terms of delay and energy consumption. This trade-off highlights the importance of carefully considering the initialization strategy for optimizing overall training efficiency.

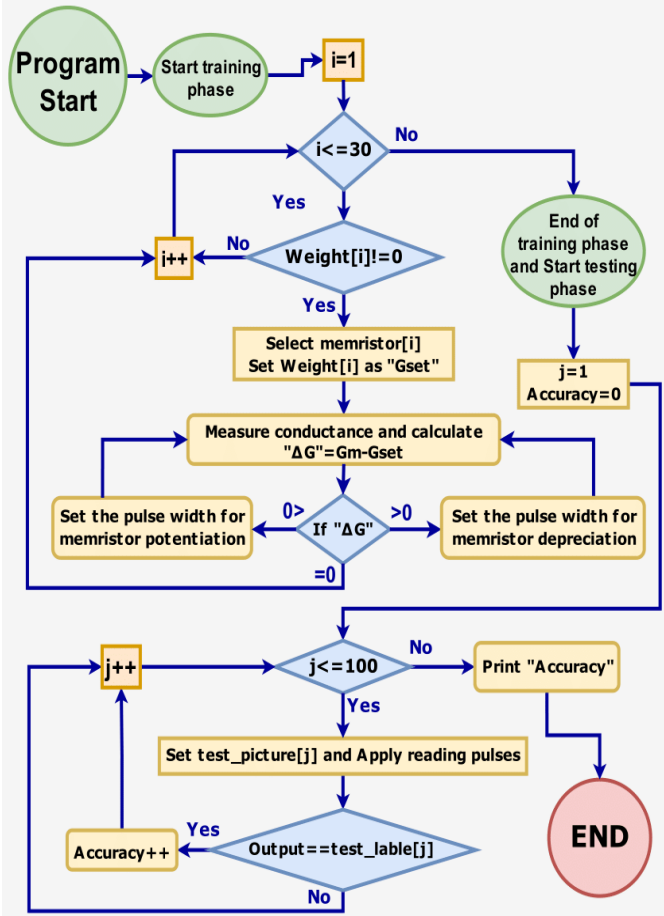

**Fig. 6.** Flowchart of the on-chip training and inference algorithm for the memristor-based neural network.

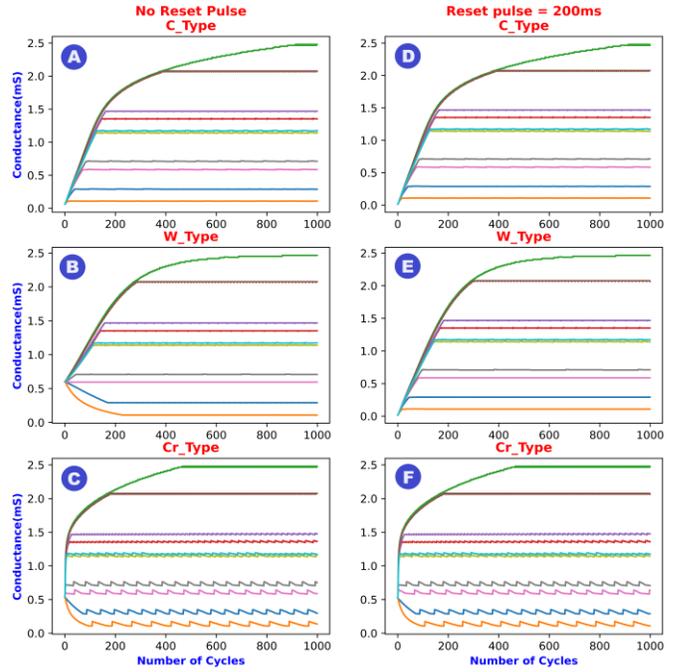

**Fig. 7.** Comparison of training cycles required to achieve target conductance levels for all memristors with non-zero weights. Graphs (A), (B), and (C) show the results without an initial reset pulse, while graphs (D), (E), and (F) illustrate the results with a 200 ms initial reset pulse. The comparison is performed for three memristor models: carbon (A and D), tungsten (B and E), and chromium (C and F). Note that due to its inherent characteristics, the chromium memristor exhibits an initial conductance jump to 0.5 mS even after the reset pulse, as further discussed in Section VI.

After resetting the memristor, the algorithm measures its conductance and calculates the difference (ΔG) between the measured value and the desired weight. Based on ΔG, the algorithm determines the pulse width and polarity for the next training pulse. This "training cycle" repeats, with ΔG recalculated after each pulse until the memristor reaches the target conductance. This process continues for all memristors with non-zero weights.

Following training, the inference phase begins. As described in Section IV, test data is applied to the network inputs over 100 cycles. The network's accuracy is evaluated by comparing the output neuron's state with the correct label for each input image. After these 100 cycles, the overall accuracy of the network is determined.

## VI. RESULTS

### A. Comparison of Network Accuracy in Numerical and Circuit Implementations and Device-to-Device Variation Effect

Addressing D2D variation in memristor-based neural networks is crucial for real-world applications [46], as these variations can lead to computational inaccuracies and reduced learning efficiency. To assess this impact, we compared circuit simulations under two conditions: one using an ideal memristor model for all network memristors, and the other randomly distributing three different memristors in the NN circuits to simulate D2D variations (Fig. 8). These results were also compared with the outcomes of a numerical implementation using the "Keras" library to show the degradation due to on-chip training procedure.

For each training session, a dataset of 1,000 samples, with varying levels of noise (5%, 10%, 15%, and 20%) at neural network inputs, representing pixel malfunctions. To investigate whether incorporating noisy training data improves the neural network's robustness to noisy input during testing, we conducted a four-phase training and evaluation process (Subplots in Fig. 8). After each training phase, test datasets with four noise levels were applied resulting in 16 distinct values of accuracy.

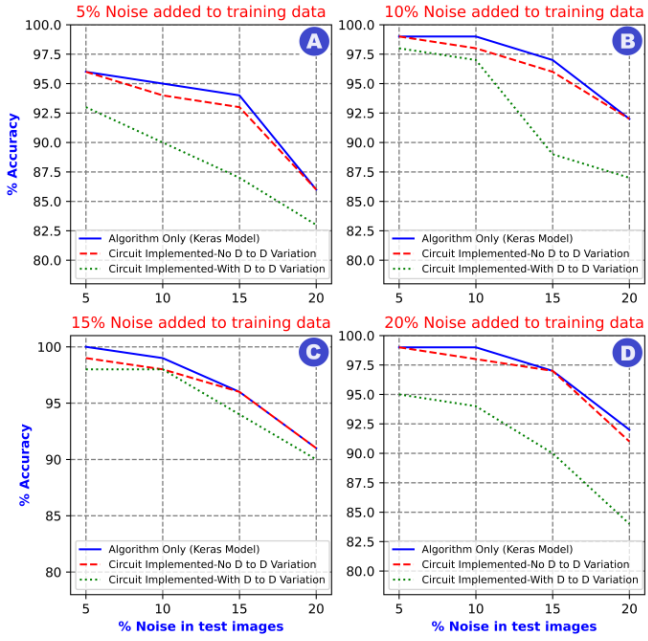

**Fig. 8.** Comparison of classification accuracy for the proposed neural network across software implementation, ideal hardware implementation, and hardware implementation with device-to-device variations. The network was trained and tested with varying levels of noise (representing pixel malfunctions) in the input images to assess the impact of noise and device variability on performance.

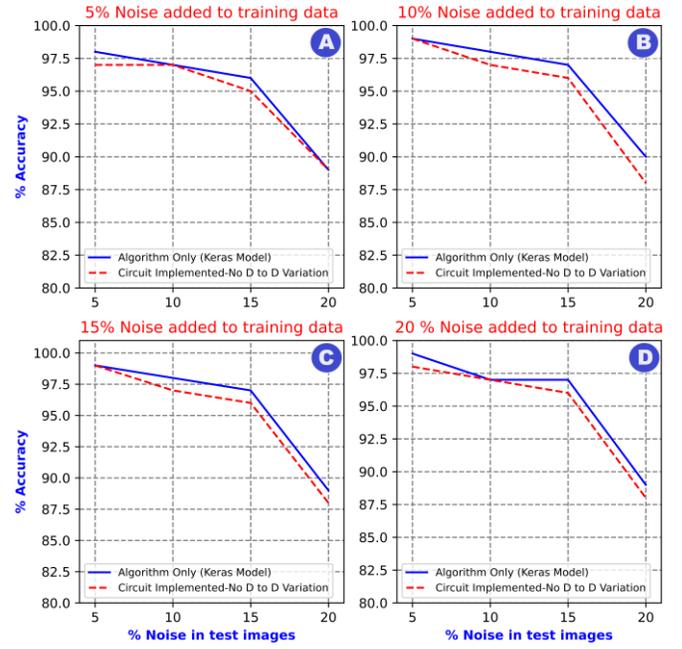

**Fig. 9.** Results of the retraining of the simulated neural network for classification between the characters 'T' and 'H'. This figure illustrates the network's capability to perform binary classification between various pairs of 9-pixel images.

First, the figures reveal that the software-based implementation (blue curves) achieved higher accuracy, while the hardware simulations provided more realistic, albeit lower, accuracy values. On-chip training of memristors resulted in up to ~2% accuracy reduction in the worst cases. However, incorporating D2D variations further degraded accuracy, reaching up to a 9% reduction in some cases compared to hardware implementations without variations. These findings highlight the critical need for efficient on-chip training circuits with minimal degradation in on-chip designs and underscore the importance of developing novel strategies to mitigate the impact of D2D variations on memristor-based neural network performance.

Another outcome from Fig. 8 is that the network's accuracy is influenced by the level of noise in the training data. When trained with less noisy data, the network exhibits lower accuracy when tested with noisier inputs, and vice versa. Incorporating noisy training data, in other words, simulating potential pixel malfunctions, enhances the network's robustness to such noise during inference. However, exceeding a certain noise threshold in the training data can diminish the network's ability to generalize effectively. Without D2D variations, the network achieved exceptional accuracy, reaching 97.5% when trained with 15% noise and tested with 5% noise. Conversely, even in the most challenging scenario (trained with 5% noise and tested with 20% noise), it achieved 87% accuracy. Introducing D2D variations, by assigning different memristor models randomly, yielded a best-case accuracy of 97% (trained with 15% noise, tested with 5% noise) and a worst-case accuracy of 84% (trained with 5% noise, tested with 20% noise).

*B. Network Capability to Classify All Possible 3×3-Pixel Images*

While the previous section focused on classifying 'X' and 'O' characters, our neural network is not limited to these specific patterns. To demonstrate its generalizability to any binary classification task involving 9-pixel images, we conducted additional training and testing with the characters 'T' and 'H' (Fig. 9). The results reveal that on-chip training of memristors led to a maximum accuracy reduction of approximately 2% in the worst cases. Furthermore, increasing the noise level in the test data consistently resulted in similar accuracy reductions, regardless of the noise intensity in the training data. This observation highlights the network's consistent performance across different classification tasks and varying levels of input noise.

*C. Impact of Memristor Conductance Error on Accuracy*

Memristor conductance errors, often arising from device variability and instability, can significantly impact the accuracy and reliability of neural network computations. These errors, caused by factors like atomic-scale defects and ion migration, can lead to unpredictable changes in synaptic weights, impacting both learning and inference. Memristor conductance drift reduces the learning accuracy and the number of cycles required for convergence, thereby impacting the overall efficiency and energy consumption of the system. These issues pose open questions for the field of memristor-based neural networks, prompting ongoing research into novel materials, device architectures, and robust error correction techniques.

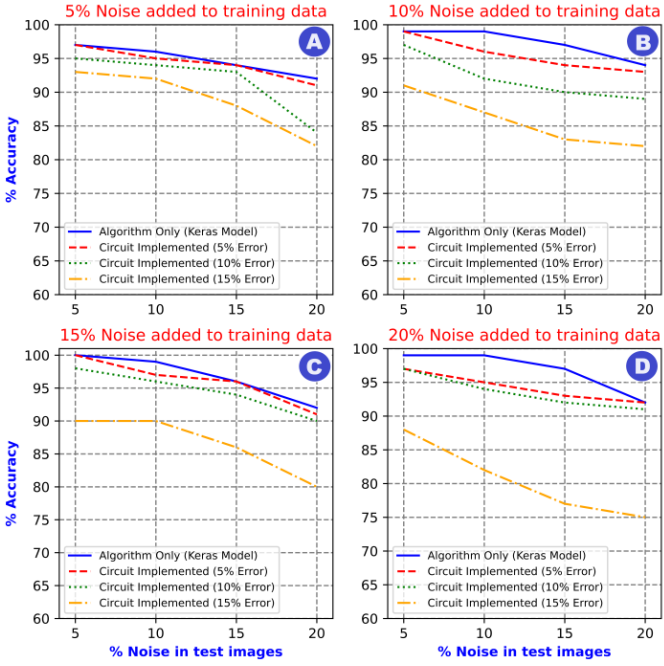

**Fig. 10.** Impact of memristor conductance error on classification accuracy. The network was trained and tested with varying levels of noise in the input images, and the allowable percentage difference between the desired and actual memristor conductance values was adjusted to simulate conductance errors. The results demonstrate the network's robustness to conductance errors up to 10%, particularly when trained with moderate noise levels. Higher conductance errors (15%) lead to a significant decline in accuracy, especially for highly noisy training data.

To examine the impact of conductance error on our network's accuracy, we modified our training algorithm to allow for larger percentage differences ($\Delta G$=0.05, 0.1, and 0.15) between the desired and actual memristor conductance values. Fig. 10 presents the results of this analysis. A conductance error of 5% or 10% does not significantly impact accuracy (~1-2%), particularly when the network is trained with 20% image noise, demonstrating robustness against these errors. However, a 15% error leads to a marked decline in accuracy, reaching up to 17% when both training and test data contain 20% image noise. This suggests that training with moderate noise levels can enhance the network's robustness to conductance errors when classifying noisy images, potentially by encouraging the network to learn more generalized features that are less sensitive to small variations in synaptic weights.

However, high noise levels in the training data can increase the network's sensitivity to these errors, likely because the network begins to overfit to the noise in the training set, resulting in a model that is less able to generalize to new, unseen data with different noise patterns. This highlights a practical limitation in incorporating noise to enhance classification robustness in memristor-based neural network design: there is an optimal noise level beyond which performance degrades.

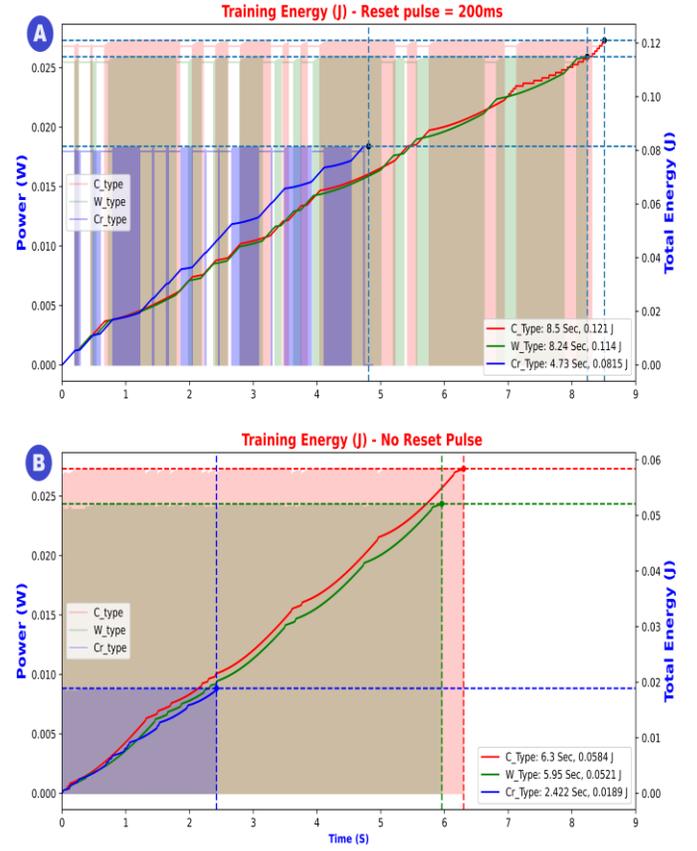

**Fig. 11.** Time, power, and energy efficiency analysis across three distinct phases. The vertical axes on the left represent the instantaneous power of the circuit, while the vertical axes on the right indicate the energy consumption of the circuit. In each phase, a model of one of the experimental memristors from Section II was assigned to all 30 memristors, followed by circuit simulation and testing. The figure also illustrates the impact of applying an initial 200ms reset pulse in the training phase, on energy consumption, and training time. Simulations were conducted for each phase with and without the reset pulse, as discussed in Section V, to comprehensively assess the performance variations.

Despite this limitation, our proposed on-chip training system allows for adjustable retraining, enabling performance restoration in case of accuracy degradation due to conductance errors or shifts in the noise characteristics of the input data.

*D. Time and Energy Consumption*

Memristor-based neural networks offer significant potential for reducing the time and energy required for data processing by enabling in-memory computing. Their non-volatile nature and ability to co-locate storage and computation allow for highly parallel and energy-efficient matrix-vector multiplications, the fundamental operations in neural networks, leading to faster training convergence and lower power consumption. These advantages make memristor-based neural networks a promising alternative for energy-efficient AI systems.

In this study, we analyze the time and energy efficiency of our memristor-based neural network across three distinct memristor models (tungsten, chromium, and carbon) by simulating the circuit's performance with and without a 200ms initial reset pulse during training, as shown in Fig. 11.

Our findings reveal that, although resetting memristors to a known baseline state can reduce the number of training cycles required to achieve the target conductance levels, the time and energy overhead associated with applying the initial reset pulse outweighs this benefit, resulting in a net increase in both training duration and energy consumption.

It can be observed that the network utilizing chromium-type memristors consistently exhibits lower time and energy consumption compared to the other types, regardless of the reset pulse application. This difference arises from the inherent physical characteristics of the chromium memristor model. Specifically, the chromium memristor exhibits a higher threshold voltage for switching from an 'HRS' to an 'LRS', indicated by a larger 'Von' parameter (Appendix D). This higher threshold voltage results in an initial conductance jump to 0.5 mS even after the application of the reset pulse, effectively providing a higher starting conductance for the training process (Figures 7C and 7F). Since the majority of target weights in this study are above 0.5 mS, the training algorithm needs to potentiate the memristor less, leading to fewer training cycles and therefore reduced time and energy consumption compared to the carbon and tungsten models, which start from a lower initial conductance.

For instance, training a carbon-type memristor to 2.5 mS requires approximately twice the number of cycles compared to a chromium-type memristor (Fig. 7).

This difference arises from the inherent characteristics of the chromium memristors. As shown in Figures, even with the initial reset pulse, the chromium memristor exhibits a conductance jump to 0.5 mS at the beginning of training due to its high 'Von' parameter (Appendix D). Since most target weights exceed 0.5 mS, the training circuit can leverage this initial conductance, significantly reducing the required training cycles.

Quantitatively, with the initial reset pulse (Fig. 11A), the total training time and energy consumption are 8.5s and 121 mJ for carbon, 8.24s and 114 mJ for tungsten, and 4.73s and 81.5 mJ for chromium, respectively. Without the reset pulse (Fig. 11B), these values decrease to 6.3s and 58.4 mJ for carbon, 5.95s and 52.1 mJ for tungsten, and 2.422s and 18.9 mJ for chromium. Additionally, the energy consumption per classification task during inference was measured as 0.115 fJ for carbon, 0.117 fJ for tungsten, and 0.111 fJ for chromium.

## VII. Conclusion

This paper presented the design, implementation, and evaluation of a memristor-based neural network hardware accelerator tailored for on-chip training and inference. We integrated realistic memristors to explore the true potential and challenges of on-chip training and inference, accounting for device non-idealities often overlooked in idealized simulations.

Our findings demonstrate the feasibility of achieving high classification accuracy with a memristor-based neural network. We highlighted the importance of incorporating moderate noise levels during training to enhance the robustness of both device variations and noisy input data. The analysis of time and energy consumption revealed efficiency advantages, particularly for the chromium memristor model, underscoring the potential of memristor-based accelerators for low-power edge computing applications. Furthermore, our study revealed critical considerations for practical implementation, such as the trade-offs associated with memristor initialization and the impact of conductance errors on accuracy. While this study focused on a simplified memristor-based neural network due to computational constraints, our proposed hardware architecture and training algorithm are inherently scalable to larger networks. Future work will focus on extending this approach to larger network architectures and more complex image classification tasks, further optimizing the on-chip training algorithm, and exploring the use of more complex training algorithms and error correction techniques to mitigate the impact of device non-idealities.

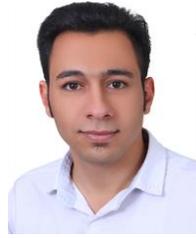

**Mohammad Reza Eslami** was born in Isfahan, Iran, in 1991. He received his bachelor's degree in Electronics-Engineering from the ACECR Institute of Higher Education (Isfahan Branch), Iran, in 2014 and his M.S. degree in Digital-Electronic-Systems in 2022 from Shahid Beheshti University in Tehran (SBU), Iran. His research interests are in the field of designing and modeling microelectromechanical circuits M/NEMS (including sensors and logic circuits), and neuromorphic circuits for the hardware implementation of artificial neural networks for energy-efficient and edge applications.

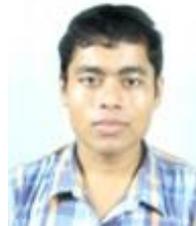

**Dhiman Biswas** was born in the Indian state of West Bengal on 13 th September 1998. He received his Bachelor's degree in Science (BSc) with honors in Physics in 2019 from the University of Calcutta in India and his master's degree in science (MSc) in physics from the Indian Institute of Technology-Guwahati (IITG) in 2021. He is currently a graduate student at the University of Oklahoma under the supervision of Professor Venky Venkatesan and Yaser Banadaki. He has experience in simulating nanoplasmonic devices and open quantum systems. His current interest in research is Neuromorphic Computing, specifically the fabrication and characterization of memristive devices.

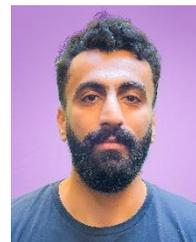

**Soheib Takhtardeshir** was born in Sanandaj, Iran, in 1991. After obtaining his B.Sc. in Electrical Engineering in 2013 and M.Sc. in Digital Electronic Engineering in 2021 from Shahid Beheshti University, Tehran, he began his Ph.D. studies in 2022 through a collaboration between Mid Sweden University and INRIA, Rennes, France, as part of the European Joint Doctoral Programme on Plenoptic Imaging (PLENOPTIMA). His research endeavors are focused on Light Field imaging, specifically its processing, representation, and compression, showcasing his commitment to advancing digital electronic engineering and plenoptic imaging technologies.

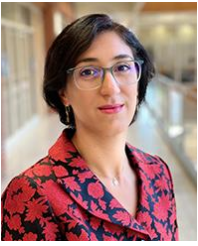 **Sarah S. Sharif** (Senior Member, IEEE), is an assistant professor in the School of Electrical and Computer Engineering at the University of Oklahoma (OU). Before joining OU, she was a postdoctoral research associate at the University of Illinois Urbana-Champaign and the University of Maryland, College Park. She holds a Ph.D. in Electrical Engineering with a minor in Physics and two M.Sc. degrees in Natural Science (Physics) and Electrical Engineering. With over 10 years of industrial experience as a research and development engineer, Dr. Sharif brings valuable expertise to her academic pursuits. From 2018 to 2020, she was an active member of the LIGO group. She leads the Quantum Nanophotonic Engineering Technology & System (QNETS) group at OU, focusing on the development of next-generation computing and sensing technologies through optical, quantum optical, and neuromorphic devices and systems. *Email:* s.sh@ou.edu

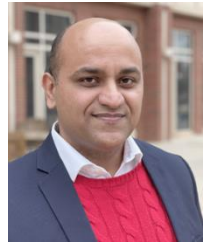 **Yaser M. Banad** (Senior Member, IEEE), is an Associate Professor at the School of Electrical and Computer Engineering, University of Oklahoma, and holds M.Sc. and Ph.D. degrees in electrical and computer engineering from Louisiana State University (2016). An author of over 100 peer-reviewed publications, Dr. Banad's research spans neuromorphic computing, energy-efficient devices, and circuits design, neural-inspired artificial intelligence acceleration, and material analysis for semiconductor technologies. He directs the Neuromorphic Intelligent Computing Systems (NICS) lab at OU, dedicated to advancing reliable, energy-efficient neuromorphic engineering from materials and devices to systems, algorithms, and applications. *Email:* bana@ou.edu